\documentclass{article}

\usepackage{microtype}
\usepackage{graphicx}
\usepackage{booktabs} 

\usepackage{amsmath}
\usepackage{amsfonts}
\usepackage{subfig}
\usepackage{multirow}
\usepackage{siunitx}
\usepackage{makecell}
\usepackage{xcolor}

\usepackage[table]{xcolor}
\definecolor{lightpink}{RGB}{255,230,240}

\newcommand{\CMT}[1]{{\it \color{red} comment: #1}}

\newcommand{\SYS}{Kitty}

\usepackage{hyperref}

\usepackage[accepted]{mlsys2025}

\mlsystitlerunning{\SYS{}: Accurate and Efficient 2-bit KV Cache Quantization with Dynamic Channel-wise Precision Boost}

\begin{document}

\twocolumn[
\mlsystitle{\SYS{}: Accurate and Efficient 2-bit KV Cache Quantization with Dynamic Channel-wise Precision Boost}



\mlsyssetsymbol{equal}{*}

\begin{mlsysauthorlist}
\mlsysauthor{Haojun Xia}{equal,usyd}
\mlsysauthor{Xiaoxia Wu}{equal,together}
\mlsysauthor{Jisen Li}{equal,uiuc}
\mlsysauthor{Robert Wu}{together}
\mlsysauthor{Junxiong Wang}{together}
\mlsysauthor{Jue Wang}{together}
\mlsysauthor{Chenxi Li}{together}
\mlsysauthor{Aman Singhal}{together}
\mlsysauthor{Alay Dilipbhai Shah}{together}
\mlsysauthor{Alpay Ariyak}{together}
\mlsysauthor{Donglin Zhuang}{usyd}
\mlsysauthor{Zhongzhu Zhou}{usyd}
\mlsysauthor{Ben Athiwaratkun}{together}
\mlsysauthor{Zhen Zheng}{msft}
\mlsysauthor{Shuaiwen Leon Song}{together}
\end{mlsysauthorlist}

\mlsysaffiliation{usyd}{University of Sydney}
\mlsysaffiliation{together}{Together AI}
\mlsysaffiliation{uiuc}{University of Illinois Urbana-Champaign}
\mlsysaffiliation{msft}{Microsoft}

\mlsyscorrespondingauthor{Shuaiwen Leon Song}{leon@together.ai}
\mlsyscorrespondingauthor{Zhen Zheng}{zhengzhen@microsoft.com}

\mlsyskeywords{Machine Learning, MLSys}

\vskip 0.3in

\begin{abstract}
The KV cache is a dominant memory bottleneck for LLM inference. While 4-bit KV quantization preserves accuracy, 2-bit often degrades it, especially on long-context reasoning. We close this gap via an algorithm–system co-design for mixed-precision KV caching: \emph{\SYS{}}. On the algorithm side, extensive experiments show that \emph{Dynamic Channel-wise Precision Boost} — which ranks Key-cache channels by sensitivity and keeps only a small fraction at higher precision — maintains near-zero loss in accuracy drop while approaching 2-bit memory. 
The main challenge is handling dynamic 4-bit channel boosts while keeping the page layout coalesced and the dequantization uniform, with no scattered reads or hard-coded masks. \emph{\SYS{}} addresses these issues by decompose each
mixed-precision Key page into two tensors with unified
2-bit precision. Based on this, Kitty provides a page-centric KV layout, Triton-compatible page dequantization kernels, and a lightweight runtime pipeline that preserves coalescing and avoids divergence. Across seven tasks and two model families (Qwen3, LLaMA3), \emph{\SYS{}} cuts KV memory by nearly $8\times$ with negligible accuracy loss, enabling up to $8\times$ larger batches and $2.1\times$–$4.1\times$ higher throughput under the same memory budget.
We release the full implementation of Kitty at \url{https://github.com/Summer-Summer/Kitty}.
\end{abstract}
]



\printAffiliationsAndNotice{\mlsysEqualContribution} 

\section{Introduction}

As large language models (LLMs) advance, their ability to process extremely long contexts (e.g., 128K tokens~\citep{GPT4, LLaMA3}) has enabled powerful applications such as detailed document understanding, extended dialogues, and complex reasoning (e.g., chain-of-thought~\citep{ChainOfThought}).
However, their progression has exposed a severe systems bottleneck: the enormous memory footprint of the KV cache.
Unlike model weights (which are static), the size of the KV cache grows proportionally with both context length and batch size, leading to outsized GPU memory footprints in long-context inference.
For a model like LLaMA3-70B~\citep{LLaMA3}, serving 32 requests with a 128K sequence requires more than 1.2~TB of KV cache storage, which is nearly an order of magnitude larger than the model weights themselves.  
But state-of-the-art data-center GPUs such as the NVIDIA B200 GPUs provide only 192~GB of memory and cost over \$30{,}000 each, making such deployments prohibitively expensive.
The massive KV cache further exacerbates inference latency due to the heavy data movement between GPU memory and compute units.

KV cache quantization offers a compelling direction to address this gap, especially the post-training scheme which can be applied without re-training or fine-tuning.  
Moreover, with proper system support (e.g., \citet{BitDecoding}), it can effectively reduce both memory consumption and inference latency.  
Unlike token pruning, quantization preserves all contextual information without discarding tokens.  
Nevertheless, low-bit (e.g. 2-bit) KV cache quantization remains challenging in practice.  
Our empirical investigation shows that quantizing the KV cache to 4-bit precision using the state-of-the-art method KIVI~\citep{KIVI} can maintain accuracy.  
However, further reducing to 2 bits significantly harms model accuracy across a range of downstream tasks.  
These findings suggest that more nuanced approaches are required to unlock the full potential of KV compression at even lower precision.

\begin{figure}[t]
  \centering
  \includegraphics[width=0.98\linewidth]{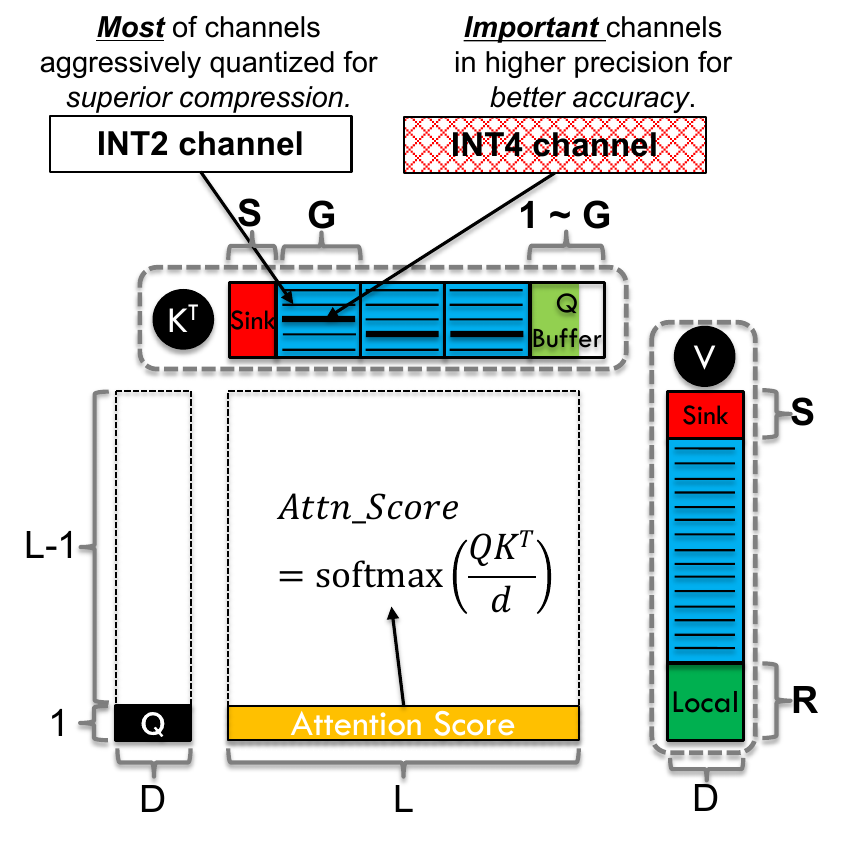}
  \caption{
    Illustration of our KV cache quantization scheme, \textbf{\SYS{}}.  
    The figure shows the decoding phase, where only the current \emph{query} vector (a single token) participates in computation, 
    while all previously stored \emph{key} and \emph{value} vectors in the KV cache are reused.  
    The key cache is organized into three parts:
    (i) \emph{Sink} (initial tokens kept in FP16),  
    (ii) a \emph{Q-Buffer} (quantization buffer, temporarily storing the FP16 KVs before forming a quantization group), and  
    (iii) \emph{quantized channels} (Most channels quantized to INT2 for maximum compression, a small fraction of channels preserved in INT4 for accuracy preservation).
    The value cache is quantized per-token with a sliding window, where both \emph{Sink} (initial tokens) and the most recent \emph{Local} (local tokens) are retained in FP16.  
    Here, $L$ denotes the sequence length and $D$ the head size; $S$ is the number of preserved sink tokens, $G$ the quantization group size, and $R$ the local window size.
    The default configuration is: $S=32$, $R=128$, $G=128$, which provides a good balance between accuracy and memory savings.
  }
  \label{fig:algorithm_overview}
\end{figure}

To address this problem, we propose a novel approach for 2-bit KV cache quantization, which we call \emph{channel-wise precision boost}.  
The key insight is that applying a uniform precision across the entire KV cache is suboptimal.  
Orthogonal to prior work~\citep{H2O}, which preserves \textbf{important tokens} in higher precision, \emph{channel-wise precision boost} preserves only \textbf{critical channels} in higher precision while aggressively compressing the remainder.
Our method is motivated by the observation that certain \textit{critical channels} within the key cache play a disproportionately important role in maintaining model accuracy.  
Based on this novel method, we propose a systematic quantization algorithm, \emph{\SYS{}}, shown in Figure~\ref{fig:algorithm_overview}.
Furthermore, we design and build an end-to-end inference system for our novel 2-bit KV quantization algorithm on GPU platforms.
To support our novel KV cache memory layout, we developed GPU kernels with Triton~\citep{Triton} for: (1) efficient quantization of newly generated KV vectors and cache updates, and (2) efficient execution of attention mechanism.

This paper makes the following contributions:
\begin{itemize}
    \item We observe that state-of-the-art 2-bit KV cache quantization methods like KIVI~\cite{KIVI} substantially degrade the accuracy of reasoning LLMs, revealing a critical representation bottleneck for low-bit KV cache quantization.
    \item We introduce \emph{Dynamic Channel-wise Precision Boost}, a novel 2-bit quantization algorithm for KV cache, inspired by the key observations in \emph{channel-wise patterns} and \emph{channel-wise quantization sensitivity}.
    Combined with other optimizations, we propose a systematic KV cache quantization scheme called \textbf{\SYS{}}.
    Extensive experiments show that \SYS{} can significantly outperform prior works in accuracy.
    \item We propose novel system designs and build an end-to-end inference system for our novel 2-bit KV quantization algorithm on GPU platforms. This system desgin solve the channels of handling dynamic 4-bit channel boosts while keeping the page layout coalesced and the dequantization uniform, with no scattered reads or hard-coded masks. 
    Our inference system enables $8\times$ larger batch sizes and achieves $2.1\times \to 4.1\times$ higher inference throughput compared to the FP16 baseline while maintaining the same memory budget.
\end{itemize}
The remainder of this chapter is organized as follows. Section~\ref{sec:background} details the background and motivation.
Section~\ref{sec:algorithm_design} describes our algorithm design for 2-bit KV cache quantization.
We present our system designs in Section \ref{sec:system_design}.
Section~\ref{sec:evaluations} details our experimental results of inference accuracy and throughput.

\section{Background \& Motivation}
\label{sec:background}
The KV cache can help avoid the expensive recomputation of historical key and value tensors by retrieving them from cache memory, significantly accelerating inference.
However, the cache size grows linearly with context length.
As real-world applications scale, the KV cache quickly dominates memory usage and becomes a critical bottleneck for inference.

\subsection{Existing KV Quantization Strategies}
Low-bit quantization has emerged as a promising direction to reduce the memory footprint of the KV cache.
The concept is to convert full-precision keys and values (e.g., FP16) into compact representations such as 8-bit, 4-bit, or even 2-bit formats.
Unlike weight quantization, which is static and applied once after training, KV cache quantization must be performed dynamically during runtime, making the problem more challenging.

\textbf{Per-token vs. Per-channel Quantization.}
The Key or Value cache can be represented as a matrix in shape $(B, H, L, D)$, where $B$ is the inference batch size, $H$ is the number of KV head, $L$ is the sequence length, and $D$ is hidden size for each head.
Thus, for each head of a certain request, the KV cache has shape $(L, D)$, which is a 2-D dimensional matrix consisting of $L$ tokens and each token is a vector of size $D$.

As a result, the KV cache can be quantized along two different dimensions, $L$ and/or $D$.
Per-token quantization applies a separate scale (and optionally zero-point) to each token’s key or value vector, so each token can be quantized separately.  
Per-channel quantization instead computes a scale per channel, shared across all tokens, and has been observed to preserve accuracy better for Key cache quantization~\citep{KIVI, KVQuant, zhang2024kv}.  
In practice, most recent works adopt per-channel quantization for the K cache and per-token quantization for the V cache.

\textbf{Mixed-precision Quantization.}  
Quantizing the entire KV cache to a very low precision (e.g., 2-bit) typically hurts model quality severely.  
To mitigate this, hybrid schemes preserve a subset of elements in higher precision while quantizing the rest.  
For example, KVQuant~\citep{KVQuant} identifies outliers and stores them in FP16 with a sparse representation.
However, KVQuant is not hardware-friendly
and usually suffers from low system-level efficiency, since it introduces additional runtime overhead from sparse-dense multiplications, which could be slow on GPUs~\citep{FlashLLM, Sputnik}.
KIVI~\citep{KIVI} and BitDecoding~\citep{BitDecoding} retain the most recent tokens in full precision to preserve an accurate local context, but they still suffer from severe accuracy degradations in low-precision regimes (e.g.~2-bit).
MiniKV~\citep{sharma2024minikv} proposes a layer-discriminative bit allocation scheme for each layers.
KVTuner~\citep{KVTuner} proposes to tune layer-wise mixed precision bitwidths for the KV cache.
QuaRot~\citep{QuaRot} applies orthogonal rotations to activations and KV cache so that the distributions become outlier-free.

\textbf{Non-KV Quantization.}
While our focus is on KV quantization, it is worth noting that several techniques for weight and activation quantization are orthogonal and potentially complementary \cite{SmoothQuant, lin2024awq}.




\subsection{Motivation: Accuracy Drop for Low-bit KV}
\label{sec:motivation}
Despite progress, existing solutions struggle to maintain accuracy when pushing the precision of KV cache below 4 bits.
\citet{quantization_hurts_reasoning} shows that quantizing KV cache to 3 bits already causes significant accuracy degradations with state-of-the-art quantization algorithms~\citep{KVQuant, QuaRot}.
It follows that 2-bit precision should be even more challenging.

We empirically observe that while 4-bit KV cache quantization can match FP16 closely, reducing it further to 2-bit (INT2) leads to substantial accuracy degradation across reasoning and generation benchmarks.

As Table~\ref{tab:kivi_kv2} shows, reducing the KV cache precision from 16 to 4 bits causes almost no degradation in accuracy for Qwen3-8B and LLaMA3-8B.
However, further reducing the cache to 2 bits greatly deteriorates the average drops to -15.23 and -10.15 respectively.
These results highlight that simply applying existing quantization methods to KV cache is insufficient, and thus motivate a line of inquiry: \textbf{can more nuanced techniques mitigate these losses and enable practical low-bit KV cache quantization for long-context LLM inference?}

\begin{table}[th]
\centering
\caption{
Accuracy degradation of low-bit KV cache quantization on Qwen3-8B~\citep{Qwen3} and LLaMA3-8B~\citep{LLaMA3}.
While 4-bit KIVI~\citep{KIVI} maintains accuracy, 2-bit KIVI shows a significant drop.
\textit{Note:} “MATH” and “GPQA” denote the \textit{MATH-Algebra} and \textit{GPQA-Diamond} subsets.
}
\label{tab:kivi_kv2}
\begin{tabular}{llccc}
\toprule
\textbf{Model} & \textbf{Task} & \textbf{FP16} & \textbf{KIVI-4bit} & \textbf{KIVI-2bit} \\
\midrule
\multirow{6}{*}{\rotatebox[origin=c]{90}{\textbf{\makecell[c]{Qwen3\\-8B}}}}
 & GSM8K   & 94.79 & 94.41 & 89.13 \\
 & MATH    & 88.26 & 88.46 & 47.29 \\
 & GPQA    & 40.71 & 38.98 & 32.24 \\
 & HumanEval & 84.82 & 84.09 & 76.89 \\
 & AIME24  & 71.67 & 77.67 & 57.00 \\
 & AIME25  & 66.00 & 65.33 & 52.33 \\
\midrule
\multirow{4}{*}{\rotatebox[origin=c]{90}{\textbf{\makecell[c]{LLaMA3\\-8B}}}}
 & GSM8K   & 76.75 & 76.09 & 63.58 \\
 & MATH    & 47.15 & 47.85 & 31.45 \\
 & GPQA    & 26.94 & 25.82 & 23.88 \\
 & HumanEval & 63.96 & 62.07 & 55.30 \\
\bottomrule
\end{tabular}
\end{table}

\section{Algorithm Design}
\label{sec:algorithm_design}

As discussed in Section~\ref{sec:motivation}, pushing KV cache quantization to extremely low precision (e.g. INT2) to save memory comes at the cost of a wide accuracy gap.
In this section, we first present our \emph{design space exploration} in Subsection \ref{sec:design_space_exploration}, where we try to reduce this gap with straightforward optimizations.
Then, we propose \emph{channel-wise precision boost} in Subsection \ref{sec:channel_wise_precision_boost}.
We present the overall quantization scheme in Subsection \ref{sec:overall_quantization_scheme}.

\subsection{Design Space Exploration}
\label{sec:design_space_exploration}

\begin{table*}[ht]
\centering
\caption{
Accuracy comparison across KV cache schemes on Qwen3-8B~\citep{Qwen3} and LLaMA3-8B~\citep{LLaMA3}.
Rows are tasks (transposed from original tables); columns are methods.
Within each model block, the best per-row is in \textbf{bold}.
\textit{Note:} ``MATH'' = MATH-Algebra, ``GPQA'' = GPQA-Diamond.
Algorithm names ending with an asterisk (*) indicate that the initial 32 tokens are preserved in full precision.
}
\label{tab:DSE}
\renewcommand{\arraystretch}{0.95}
\setlength{\tabcolsep}{6pt}
\begin{tabular}{llccccc}
\toprule
\textbf{Model} & \textbf{Task} & \textbf{FP16} & \textbf{KIVI-K2V2} & \textbf{KIVI-K2V2*} & \textbf{KIVI-K2V4*} & \textbf{KIVI-K4V2*} \\
\midrule
\multirow{6}{*}{\textbf{\makecell[c]{Qwen3-8B}}}
 & GSM8K         & 94.79 & 89.13 & 89.71 & 90.14 & \textbf{93.96} \\
 & MATH          & 88.26 & 47.29 & 74.92 & 82.50 & \textbf{87.92} \\
 & GPQA          & 40.71 & 32.24 & 36.02 & 35.82 & \textbf{40.51} \\
 & HumanEval     & 84.82 & 76.89 & 78.54 & 81.77 & \textbf{83.41} \\
 & AIME24        & 71.67 & 57.00 & 67.67 & 71.00 & \textbf{76.00} \\
 & AIME25        & 66.00 & 52.33 & 57.67 & \textbf{64.33} & \textbf{64.33} \\
\midrule
\multirow{4}{*}{\textbf{\makecell[c]{LLaMA3-8B}}}
 & GSM8K         & 76.75 & 63.58 & 71.04 & 72.38 & \textbf{76.62} \\
 & MATH          & 47.15 & 31.45 & 44.12 & 44.09 & \textbf{48.41} \\
 & GPQA          & 26.94 & 23.88 & 23.67 & 24.80 & \textbf{25.20} \\
 & HumanEval     & 63.96 & 55.30 & 56.71 & 59.02 & \textbf{62.80} \\
\bottomrule
\end{tabular}
\end{table*}

\begin{figure}[ht]
  \centering
  \subfloat[Visualization of Key-cache.
  \label{fig:Visual_Qwen3_8B_layer10}]
  {\includegraphics[width=0.3\textwidth]{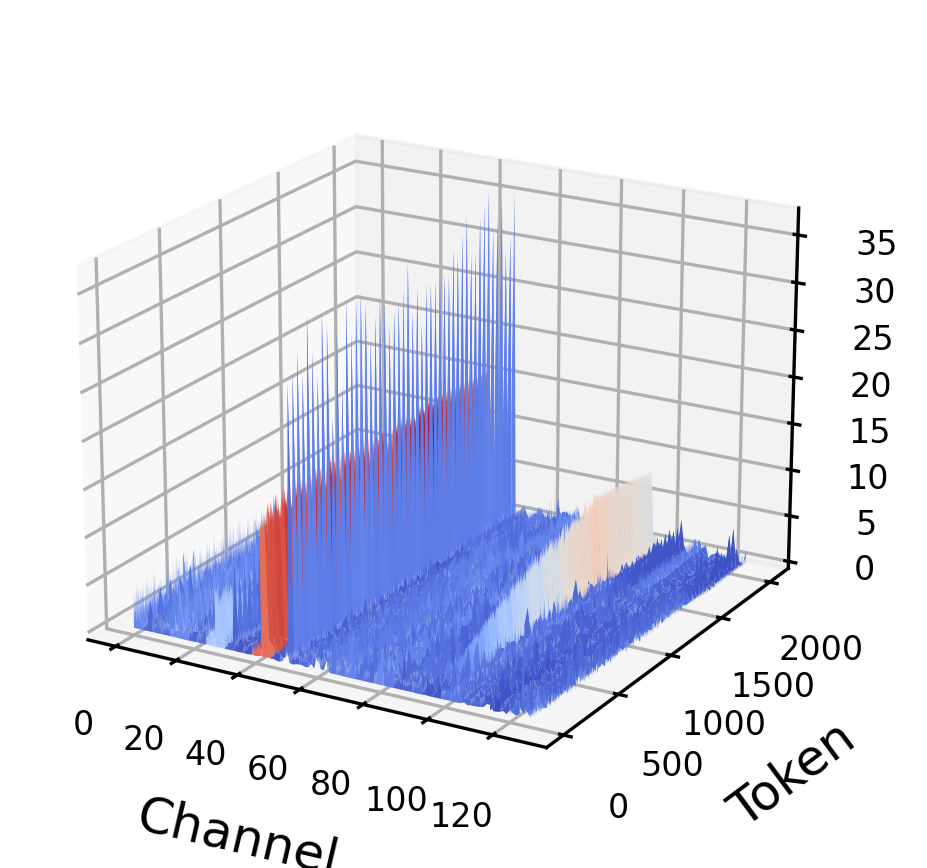}}
  \\
  \subfloat[MSE error on attention score after quantizing each channel.
  \label{fig:MSE_Qwen3_8B_layer10}]
  {\includegraphics[width=0.98\linewidth]{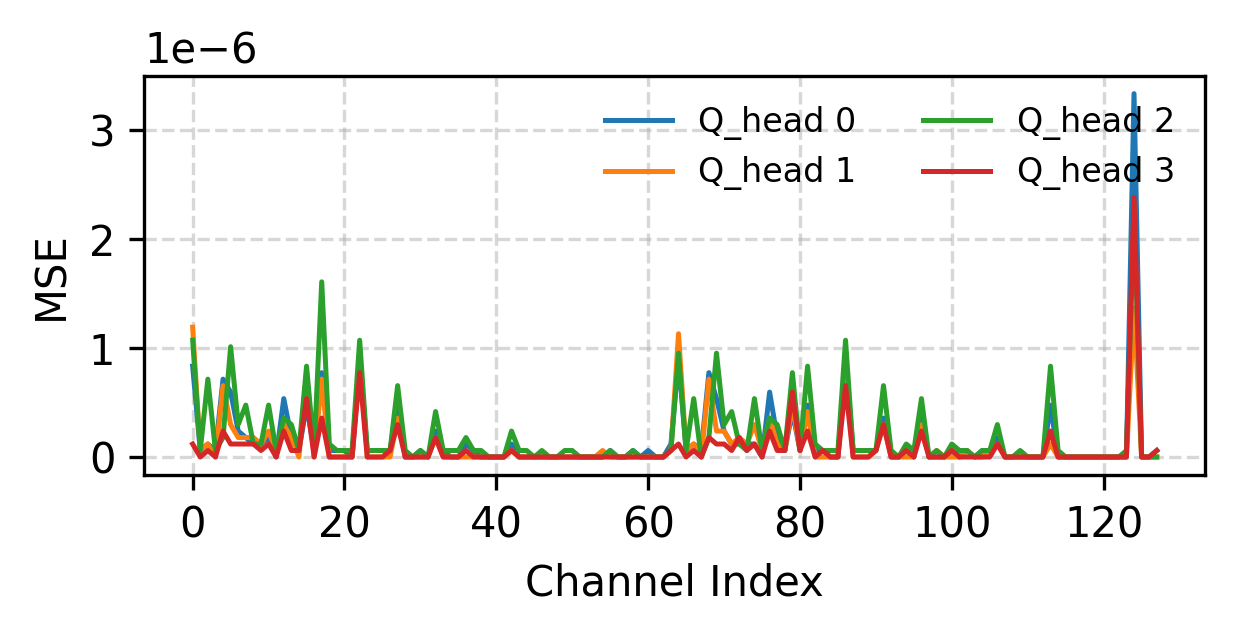}}
  \\
  \caption{
  Visual and statistical analysis of Key cache from Layer 10, Qwen3-8B.
  (a) Visualization of Key-cache magnitude from the first KV head.
  The uneven distribution, with a few channels showing consistently high magnitudes, motivates a channel-aware approach to quantization.
  The vertical axis denotes activation magnitudes, while the horizontal axis spans the \emph{token} and \emph{channel} dimensions.
  (b) The mean squared error (MSE) between the original attention score matrix with its perturbed counterpart after quantizing each channel of Key cache.
  The pattern is consistent between different Q heads who share the same Key cache due to grouped-query attention~\citep{GQA}.)
  Similar patterns are observed on other layers/models.
  }
  \label{fig:key_cache_visualization}
\end{figure}

\paragraph{Preserving the Initial Tokens in Full Precision}
The initial tokens in a sequence are sometimes referred to as \emph{attention sinks}, as noted in StreamingLLM~\citep{StreamingLLM}.
These tokens typically receive disproportionately high attention weights and thus contribute significantly to the final outputs.
In this paper, we stipulate that preserving the initial tokens in full precision can help mitigate accuracy loss with negligible overhead.
We implemented \textsc{KIVI-K2V2*}, which is an algorithmic variant of KIVI, where the first 32 tokens are additionally preserved in full precision in KV cache.
Empirical results in Table \ref{tab:DSE} show that \textsc{KIVI-K2V2*} substantially alleviates accuracy degradation across benchmarks.
On average, \textsc{KIVI-K2V2*} outperforms the original \textsc{KIVI-K2V2} by +8.28 and +5.33 on Qwen3-8B and LLaMA3-8B respectively.

\paragraph{Increasing the Precision of Key Cache.}
While preserving the precision of sink-token entries recovers some performance, this technique still leaves much to be desired when compared to the FP16 baseline.
To further reduce this gap and better understand the relative importance of the Key and Value caches, we also evaluate two design variants: \textsc{KIVI-K2V\textbf{4}*} and \textsc{KIVI-K\textbf{4}V2*}.
Results in Table \ref{tab:DSE} indicate that increasing the precision of the Key cache proves substantially more effective than doing so for the Value cache.
In particular, \textsc{KIVI-K\textbf{4}V2*} outperforms \textsc{KIVI-K2V\textbf{4}*} and achieves accuracy approaching the FP16 baseline across multiple benchmarks, underscoring the importance of increasing the precision of Key cache over the Value cache.

\subsection{Channel-wise Precision Boost}
\label{sec:channel_wise_precision_boost}

In the previous subsection (\S~\ref{sec:design_space_exploration}), we observe that raising the precision of the Key cache can effectively mitigates the accuracy degradation from KV cache quantization.
It motivates a key question: \emph{rather than boosting the precision of the entire Key cache, can we obtain comparable accuracy gains by increasing the precision of only a small subset of the Key cache?}
To explore this idea, we first perform both visual and statistical analysis of the Key-cache tensors, and then propose our key innovation according to the observations.

\textbf{Observation-1: Channel-wise Patterns in Key Cache.}
We visualize the absolute values of Key-cache activations across multiple layers on Qwen3-8B, and find that the magnitudes of different channels in Key cache vary considerably.
As shown in Figure~\ref{fig:Visual_Qwen3_8B_layer10}, a subset of channels consistently exhibits higher magnitudes.
While this figure only presents the visualization of layer 10, similar pattern is observed on other layers.
These observations align with prior work~\citep{KIVI,KVQuant}, which leveraged per-channel quantization to reduce quantization errors.

These observations suggest that not all channels in the Key cache behave equally.  
Some channels appear to exert a stronger influence, while others may have only a marginal effect.  
This uneven contribution motivates a channel-aware perspective: by selectively allocating higher precision to the most influential channels, it may be possible to preserve model accuracy while still reducing the overall KV cache memory footprint.

\textbf{Observation-2: Channel-wise Quantization Sensitivity.}
To understand the importance of each channel, we isolate the impact of quantization on each individual channels.
Specifically, we quantize a single channel to 2 bits (INT2) each time while keeping other channels unchanged, and calculate its impact on the attention score:
\begin{equation}
    \text{attn\_score} = \text{softmax}\!\left(\frac{Q K^{\top}}{\sqrt{d}}\right).
    \label{equ:attn_score}
\end{equation}
We compute the \textbf{mean squared error (MSE)} between the original attention score matrix with its perturbed counterpart (after quantizing a channel), and use this MSE as the metric to quantify the sensitivity of each channel in Figure~\ref{fig:MSE_Qwen3_8B_layer10}.
Since Grouped-Query Attention (GQA)~\citep{GQA} is used in these models, where one key head is shared across multiple query heads, we report the MSE for each query head separately with different colors.

As shown in Figure~\ref{fig:MSE_Qwen3_8B_layer10}, different channels exhibit vastly different levels of sensitivity to quantization.
A small fraction of channels consistently introduce larger errors to the attention score upon 2-bit quantization, suggesting they should be preserved in higher precision, e.g. 4-bit.
Meanwhile, quantizing other channels to 2-bit causes less significant error on the attention score, indicating that they can be quantized more aggressively to 2-bit safely.
Similar patterns are observed on other layers and models.
This disparity highlights a clear opportunity:
rather than uniformly quantizing all channels to 2-bit, selectively preserving a small fraction of sensitive channels in higher precision is promising to preserve inference accuracy while reducing the size of KV cache. 
This insight forms the foundation of the following key innovation.

\paragraph{Key Innovation: Boosting Precision of Critical Channels}

Inspired by prior observations, we propose our \emph{channel-wise precision boost}, where the critical channels are preserved in higher precision (INT4) while the others are aggressively quantized to lower precision (INT2), to reduce the distortion introduced by uniform 2-bit quantization and improving inference accuracy.

To apply our \emph{channel-wise precision boost} method, we need to first identify a subset of channels to boost.  
Directly computing the sensitivity of every channel at runtime, as we did in Figures~\ref{fig:MSE_Qwen3_8B_layer10}, is prohibitively expensive which requires computing the attention score matrix and measuring its MSE against an FP16 baseline for each channel.
To address this, we approximate channel importance using lightweight heuristics that can be computed in a single pass.
In this work, we use \emph{magnitude-based selection} heuristic to approximate the channel importance. 
This heuristic is motivated by two intuitions.  
First, channels with larger average magnitudes are more susceptible to quantization error.  
Second, such channels tend to exert greater influence on attention scores.  
We therefore define the importance score as the average magnitude of channel $i$ across all tokens:
\begin{equation}
    s_i = \frac{1}{T} \sum_{t=1}^{T} \left| x_{i,t} \right|.
\end{equation} 

Note that the score of each channel is required to be computed during inference runtime.
Based on these channel-wise scores, top-K~\footnote{K is determined by the anticipated memory budget, e.g. if you want to boost 25\% of the channels to higher precision and the size per head is 128, then K is 32.} channels are selected and stored to KV cache in higher precision (e.g., INT4).
We evaluated only one heuristic and one baseline for channel selection. 
While sufficient to demonstrate the effectiveness of \emph{channel-wise precision boost} and the necessity of heuristic-guided selection over random choice, more principled or adaptive strategies may yield stronger robustness accuracy recovery. 
Exploring such strategies is a natural next step and we leave it as a future work.

\subsection{Kitty: Overall Quantization Scheme}
\label{sec:overall_quantization_scheme}

Our \textbf{\SYS{}} quantization algorithm builds upon the foundation of \textsc{KIVI}~\citep{KIVI} while introducing two key enhancements to further improve accuracy under aggressive low-bit settings.
(1) We preserve the initial tokens (\emph{Sink tokens}) in full precision for both the Key and Value caches.
(2) More importantly, our novel \emph{channel-wise precision boost} mechanism is applied to our \SYS{}.
Figure~\ref{fig:algorithm_overview} illustrates the overall design of our quantization scheme, \SYS{}, which integrates all the above optimizations into a unified scheme.

To evaluate the accuracy impact of this quantization algorithm, we integrated \emph{\SYS{}} quantization scheme into the HuggingFace Transformers framework~\citep{hf_transformers}, and built an accuracy simulation framework
to validate the accuracy impact of \SYS{}.

\section{System Design and Implementation}
\label{sec:system_design}

This section describes the system-level innovations that enable \SYS{} to efficiently execute mixed-precision quantization on GPUs.
Our design includes:
(1) a page-centric memory layout for \SYS{} KV Cache,
(2) Triton compatible GPU kernel design for dequantization, and 
(3) a lightweight runtime pipeline for \SYS{} attention execution.

\subsection{Page-Centric Memory Layout.}

To support our proposed \SYS{} algorithm, we design a memory- and compute-efficient layout for the KV cache.
Given that KV paging~\citep{PagedAttention} is essential for scalable cache management and for reducing memory fragmentation, we implement this feature by treating each quantization group (e.g., 128 tokens) as an independent page.
To enable efficient runtime loading, each Key cache page is decomposed into two 2-bit matrices: a dense matrix and a structured sparse matrix\footnote{We do not store the zero elements within the sparse matrix.}, which are stored separately.
This design simplifies the loading process, as both matrices now contain only 2-bit elements.

\begin{figure}[htb]
    \centering
    \subfloat[Memory layout of key cache.\label{fig:memory_layout_key_cache}]{
        \includegraphics[width=0.4\textwidth]{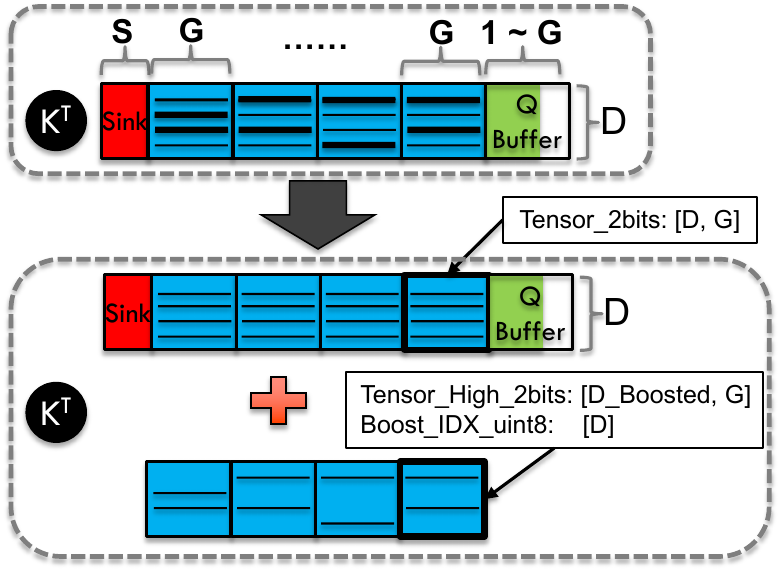}
    }
    \\
    \subfloat[Memory layout of value cache. \label{fig:memory_layout_value_cache}]{
        \includegraphics[width=0.4\textwidth]{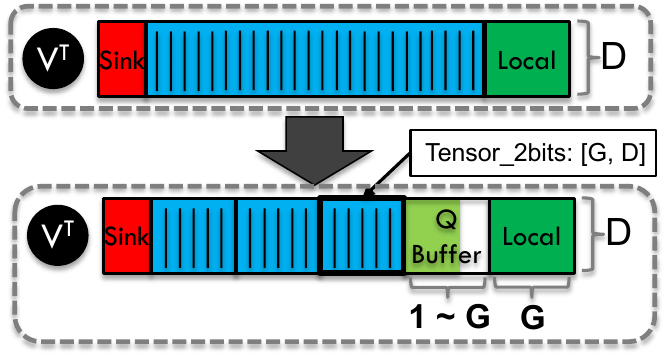}
    }
    \caption{Illustration of \SYS{}’s page-centric KV cache layout.
    }
    \label{fig:memory_layout_kv_cache}
\end{figure}

\paragraph{Paged and Quantized Key-Value Cache.}
As shown in Figure \ref{fig:memory_layout_kv_cache}, we store the KV cache with several full-precision buffers (i.e. \emph{Sink}, \emph{Q-Buffer}, \emph{Local}) in GPU memory and many fixed-size memory pages (inspired by Paged Attention~\citep{PagedAttention}).
In the design of \SYS{} shown in Figure~\ref{fig:algorithm_overview}, $G$ (e.g. 128) tokens form a quantization group in the Key cache, we naturally choose $G$ as the page size.
Thus, each page contains ($D$, $G$) elements, where D is \texttt{size\_per\_head}.
For efficient memory management, we also split the Value cache into pages.

\paragraph{Dynamic Dense-Sparse Decomposition of Quantized Pages.}
During attention computation, all pages of the KV cache must be fetched from HBM.
However, in our \SYS{} algorithm, different channels in the Key cache are quantized with mixed precision (2-bit or 4-bit), which complicates the page loading process.
To address this challenge, inspired by \citet{zeroquant(4+2)} and \citet{QuantLLM}, we decompose each mixed-precision Key page into two tensors with unified 2-bit precision, as illustrated in Figure~\ref{fig:memory_layout_key_cache}.

We store all 2-bit channels and the lower two bits of boosted 4-bit channels in a dense tensor, \texttt{Tensor\_2bits}, of shape $(D, G)$.
As shown in the lower part of Figure~\ref{fig:memory_layout_key_cache}, the higher two bits of boosted channels form a 2-bit sparse tensor.
For memory efficiency, we compactly store only the nonzero channels of this sparse tensor, resulting in \texttt{Tensor\_High\_2bits} of shape $(D_{\text{boost}}, G)$, where $D_{\text{boost}}$ denotes the number of channels promoted to 4-bit precision.

To reconstruct the full tensor, we maintain an index tensor \texttt{Boost\_IDX\_uint8} with shape $(D,)$, defining a mapping ${1, 2, \dots, D} \rightarrow {1, 2, \dots, D_{\text{boost}}}$ that maps each logical channel index to its physical offset in GPU memory.
For channels not boosted to INT4 precision, the corresponding index is set to a sentinel value of $D_{\text{boost}} + 1$.

\subsection{GPU Kernel Design for Page Dequantization.}
Reconstructing an FP16 page from our quantized representation is non-trivial, particularly for the Key cache where each channel may have a different precision (2-bit or 4-bit).  
To address this challenge, our \SYS{} system leverages the previously introduced \emph{Dense–Sparse Decomposition} and the index tensor \texttt{Boost\_IDX\_uint8} to enable fully parallel page reconstruction on GPU.

Algorithm~\ref{alg:dequant_keycache} presents the Triton-style pseudo code of this dequantization kernel, which reconstructs an FP16 tensor directly on-chip from quantized memory pages (Figure~\ref{fig:memory_layout_key_cache}).
In line~8, an \texttt{uint8} tensor of shape $(D, T)$ is loaded from HBM, where each byte encodes four consecutive 2-bit values.  
As a result, the same byte is logically reused by four times, introducing intentional redundancy in on-chip memory. 
In line~9, the lower two bits of each packed byte are extracted via a bit-shift and mask operation.  
For the higher two bits, line~10–11 conditionally fetches additional data from HBM according to the \texttt{boost\_mask}.  
Only channels promoted to INT4 precision are read from the secondary tensor (\texttt{Tensor\_High\_2bits}), where the \texttt{boost\_idx} tensor provides direct address mapping between the dense tensor and the compact boosted subspace.
Finally, the kernel combines both low and high 2-bit components, applies per-channel scaling and zero-point correction, and reconstructs the full-precision $\mathbf{K}_\text{fp16}$ page on-chip.  
This design avoids divergent memory accesses and achieves efficient bit-level unpacking and parallel reconstruction for all channels.

\begin{algorithm}[H]
\caption{Dequantize\_KeyCache\_Page()}
\label{alg:dequant_keycache}
\begin{algorithmic}[1]
\STATE shifts = [0, 2, 4, 6, 0, ...]
\STATE \textbf{Input:} Quantized cache $\mathbf{C}$, metadata $\mathbf{M}$, boosted channel number $D_\text{boost}$
\STATE \textbf{Output:} Dequantized key page $\mathbf{K}_\text{fp16} \in \mathbb{R}^{D \times T}$
\STATE $\text{scale}, \text{zero\_point} \leftarrow \text{LoadMeta}(\mathbf{M})$
\STATE $\text{boost\_idx} \leftarrow \text{Load}(\mathbf{C})$
\STATE $\text{boost\_mask} \leftarrow (\text{boost\_idx} \leq D_\text{boost})$
\STATE \textcolor{gray}{\#$\text{X}_\textbf{low}$ is an uint8 tensor with shape (D, T).}
\STATE $\mathbf{X}_\text{low} \leftarrow \text{LoadLowBits}(\mathbf{C})$
\STATE $\mathbf{X}_\text{low} \leftarrow (\mathbf{X}_\text{low} \gg \texttt{shifts}) \,\&\, 0x3$
\STATE \textcolor{gray}{\#$\text{X}_\textbf{high}$ is an uint8 tensor with shape (D, T).}
\STATE $\mathbf{X}_\text{high} \leftarrow \text{LoadHighBits}(\mathbf{C}, \text{boost\_idx}, \text{boost\_mask})$
\STATE $\mathbf{X}_\text{high} \leftarrow (\mathbf{X}_\text{high} \gg \texttt{shifts}) \,\&\, 0x3$
\STATE \textcolor{gray}{Combining and dequantization.}
\STATE $\mathbf{X} \leftarrow \mathbf{X}_\text{low} \,|\, (\mathbf{X}_\text{high} \ll 2)$
\STATE $\mathbf{K}_\text{fp16} \leftarrow \mathbf{X} \odot \text{scale} + \text{zero\_point}$
\STATE \textbf{return} $\mathbf{K}_\text{fp16}$
\end{algorithmic}
\end{algorithm}

\subsection{Lightweight Attention Execution Pipeline.}
In this subsection, we present our novel execution pipeline for attention layers.
The execution pipeline of FP16 attention is straightforward, where the KV cache is first updated and then the attention output is computed.
However, the situation becomes more complicated for the \SYS{} KV cache, as it requires coordinated management across multiple heterogeneous memory components, including the full-precision buffers (\emph{Sink}, \emph{Q-Buffer}, and \emph{Local}) and the quantized pages stored in HBM.
Each component serves a distinct purpose in balancing precision, memory efficiency, and temporal locality.

To synchronize these heterogeneous components efficiently, \SYS{} employs a lightweight, three-stage execution pipeline.
(1) inserting the new KV vectors into a static buffer (e.g. \emph{Sink}, \emph{Q-Buffer}, or \emph{Local});
(2) loading the KV cache from our heterogeneous memory layout, reconstructing the FP16 KV cache with on-chip memory and computing the attention output;
(3) and (optionally) performing quantization and packing of the KV vectors from the full precision buffers.

\paragraph{Step (1): Inserting the New KV Vectors in Full Precision.}
The newly generated KV vectors will be inserted into the KV cache before the attention computation. There might be multiple destinations for the KV vectors depending on the current state of each buffer.
If the \emph{Sink} is not full, the new KV will be directly inserted into it.
Otherwise, the new Key tensor will be inserted into \emph{Q-Buffer}, which is guaranteed to be not full due to Step (3), for the Key cache.
For the Value cache, we also allocate a \emph{Q-Buffer} to store the tensors to be quantized (in a large granularity rather than per-decoding-step to reduce the cost).
Differently, the new Value tensor will be inserted into the \emph{Local} buffer first if the \emph{Sink} is full.
Then, the oldest token in the \emph{Local} will be evicted and inserted into \emph{Q-Buffer} if \emph{Local} is full.
This is to guarantee that there is a fixed local window for Value cache.

\paragraph{Step (2): Attention Computation.}
We implemented the overall attention computation using two customized Triton GPU kernels (\texttt{qk\_kernel} and \texttt{sv\_kernel}) and one PyTorch operator (softmax).
The attention logits are first computed with \texttt{qk\_kernel}, where it loads the Query vector and the (quantized) Key cache from GPU memory and compute the matrix multiplication.
Then, the attention score (full precision) is computed with PyTorch softmax operator.
Finally, the attention scores and the (quantized) Value cache are multiplied with \texttt{sv\_kernel}.
We leave the further optimization of these kernels (e.g., fusing these kernels into one~\citep{FlashAttention}) for future work.

\paragraph{Step (3): Quantization and Packing}
After the attention computation, quantization and packing process will be launched once the \emph{Q-Buffer} is full.
During this process, all the KV vectors in the \emph{Q-Buffer} will be quantized with our customized Triton kernel and packed into one page (described in Figure \ref{fig:memory_layout_kv_cache}).
Given that the size of \emph{Q-Buffer} is G and at most one vector can be inserted to \emph{Q-Buffer} for each decoding step, this quantization process can be launched at most once every G decoding steps.
In this way, the quantization overhead is efficiently amortized and becomes negligible.

\section{Experimental Results}
\label{sec:evaluations}

\subsection{Setup}
\textbf{Evaluated Models.}
We evaluate the accuracy of \emph{\SYS{}} on two families of open-source reasoning models: Qwen3~\citep{Qwen3} (8B, 14B, 32B) and LLaMA 3~\citep{LLaMA3} (8B, 70B).

\textbf{Evaluation Datasets.}
The accuracy impact was evaluated mainly on four reasoning benchmarks:
(1) two mathematical reasoning datasets: GSM8K~\citep{GSM8K} and MATH-Algebra~\citep{MATH};
(2) a code generation benchmark HumanEval~\citep{HumanEval}; and
(3) a graduate-level science question dataset GPQA-Diamond~\citep{GPQA}.
We also conducted extended evaluations on advanced math problems AIME24~\citep{AIME24} and AIME25~\citep{AIME25}.
For HumanEval, we report functional correctness using pass@1 as the primary metric, while for the other benchmarks we report accuracy by comparing the extracted model outputs against ground-truth labels, following standard practice.

\textbf{Evaluation Frameworks.}
We conduct end-to-end accuracy evaluation using our simulation framework.
Meanwhile, we perform system evaluations using our novel inference engine described in Section \ref{sec:system_design}.
Downstream task evaluations are performed via \emph{lm-evaluation-harness}~\citep{lm-eval}, a widely adopted framework for standardized benchmarking of large language models.  
To better capture accuracy under long-context generation, we adopt the chain-of-thought (CoT) variants of evaluation prompts. 
Following common practice, we evaluate GSM8K with 8-shot prompts, MATH with 4-shot prompts, and GPQA with 5-shot prompts.

\textbf{Evaluation Configurations.}
All evaluations are conducted using PyTorch 2.4.1 with CUDA 12.1 and FlashAttention \texttt{2.7.4.post1} \citep{FlashAttention}.
We evaluate the accuracy of Qwen3-8B and LLaMA3-8B on NVIDIA A100 GPUs \citep{Ampere_WhitePaper}, and the larger models on NVIDIA H100 GPUs \citep{Hopper_WhitePaper}.
Following common practice, we enable stochastic sampling during inference, setting the \emph{temperature} to 0.6, \emph{top\_p} to 0.95, and \emph{top\_k} to 20.
During accuracy evaluation, the maximum number of generated tokens is limited to 4096.
For AIME tasks, we extend the maximum generation length to 32768 tokens to accommodate the long-chain reasoning required by such problems.
To ensure the stability of our results, each experiment is repeated anywhere from $3$ up to $10$ times, and we report both the average accuracy and the maximum observed deviation.

\subsection{Accuracy Results: Accuracy Recovery}
\label{sec:accuracy_overview}

\begin{table*}[htb]
  \centering
  \caption{Benchmark results of different KV cache quantization methods.}
  \label{tab:accuracy_overview}
\begin{tabular}{
      l
      l
      c  
      c  
      c  
      c  
      c  
      c  
  }
    \toprule
    \multicolumn{2}{c}{\textbf{Model / Method}} 
      & \multicolumn{1}{c}{\textbf{GSM8K}}
      & \multicolumn{1}{c}{\makecell{\textbf{MATH}\\\textbf{ALGEBRA}}}
      & \multicolumn{1}{c}{\makecell{\textbf{HUMAN}\\\textbf{EVAL}}}
      & \multicolumn{1}{c}{\makecell{\textbf{GPQA}\\\textbf{DIAMOND}}}
      & \multicolumn{1}{c}{\textbf{Average}}
      & \multicolumn{1}{c}{\textbf{Drop}} \\
    \midrule
    \multirow{5}{*}{\textbf{Qwen3-8B}} 
      & \textbf{K16V16}   & \textbf{94.79{\scriptsize$\pm$}$_{0.71}$} & \textbf{88.26{\scriptsize$\pm$}$_{0.45}$} & \textbf{84.82{\scriptsize$\pm$}$_{3.72}$} & \textbf{40.71{\scriptsize$\pm$}$_{2.96}$} & \textbf{77.15} & {-} \\
      & KIVI-K2V2                  & 89.13{\scriptsize$\pm$}$_{0.51}$ & 47.29{\scriptsize$\pm$}$_{0.20}$ & 76.89{\scriptsize$\pm$}$_{3.11}$ & 32.24{\scriptsize$\pm$}$_{3.16}$ & 61.39 & -15.76 \\
      & KIVI-K2V2*           & 89.71{\scriptsize$\pm$}$_{0.63}$ & 74.92{\scriptsize$\pm$}$_{1.83}$ & 78.54{\scriptsize$\pm$}$_{2.56}$ & 36.02{\scriptsize$\pm$}$_{3.27}$ & 69.80 & -7.35 \\
     \rowcolor{lightpink} &  \SYS{}         & 93.61{\scriptsize$\pm$}$_{0.58}$ & 85.12{\scriptsize$\pm$}$_{1.54}$ & 81.77{\scriptsize$\pm$}$_{1.89}$ & 39.39{\scriptsize$\pm$}$_{4.18}$ & 74.97 & -2.18 \\
      \rowcolor{lightpink} & \SYS{}-Pro          & 94.34{\scriptsize$\pm$}$_{0.48}$ & 88.12{\scriptsize$\pm$}$_{1.26}$ & 81.34{\scriptsize$\pm$}$_{3.41}$ & 40.92{\scriptsize$\pm$}$_{5.00}$ & 76.18 & -0.97 \\

    \midrule

    \multirow{5}{*}{\textbf{Qwen3-14B}} 
      & \textbf{K16V16}   & \textbf{94.69{\scriptsize$\pm$}$_{0.45}$} & \textbf{90.68{\scriptsize$\pm$}$_{0.14}$} & \textbf{86.18{\scriptsize$\pm$}$_{2.03}$} & \textbf{47.62{\scriptsize$\pm$}$_{3.74}$} & \textbf{79.79} & {-} \\
      & KIVI-K2V2                  & 75.82{\scriptsize$\pm$}$_{1.97}$ & 83.66{\scriptsize$\pm$}$_{1.01}$ & 83.74{\scriptsize$\pm$}$_{0.41}$ & 41.50{\scriptsize$\pm$}$_{0.34}$ & 71.18 & -8.61 \\
      & KIVI-K2V2*           & 89.56{\scriptsize$\pm$}$_{0.94}$ & 83.74{\scriptsize$\pm$}$_{0.59}$ & 85.98{\scriptsize$\pm$}$_{1.83}$ & 45.24{\scriptsize$\pm$}$_{2.89}$ & 76.13 & -3.66 \\
      \rowcolor{lightpink} & \SYS{}         & 94.67{\scriptsize$\pm$}$_{0.94}$ & 90.31{\scriptsize$\pm$}$_{0.93}$ & 88.21{\scriptsize$\pm$}$_{2.24}$ & 47.11{\scriptsize$\pm$}$_{4.25}$ & 80.08 & 0.29 \\
    \midrule

    \multirow{5}{*}{\textbf{Qwen3-32B}} 
      & \textbf{K16V16}   & \textbf{91.74{\scriptsize$\pm$}$_{0.99}$} & \textbf{84.84{\scriptsize$\pm$}$_{0.93}$} & \textbf{85.98{\scriptsize$\pm$}$_{1.22}$} & \textbf{48.81{\scriptsize$\pm$}$_{1.87}$} & \textbf{77.84} & {-} \\
      & KIVI-K2V2                  & 88.17{\scriptsize$\pm$}$_{0.38}$ & 59.70{\scriptsize$\pm$}$_{0.81}$ & 84.76{\scriptsize$\pm$}$_{0.61}$ & 44.22{\scriptsize$\pm$}$_{1.19}$ & 69.21 & -8.63 \\
      & KIVI-K2V2*           & 89.31{\scriptsize$\pm$}$_{0.91}$ & 74.92{\scriptsize$\pm$}$_{0.45}$ & 85.37{\scriptsize$\pm$}$_{2.44}$ & 48.98{\scriptsize$\pm$}$_{2.55}$ & 74.65 & -3.19 \\
      \rowcolor{lightpink}& \SYS{}         & 91.56{\scriptsize$\pm$}$_{0.58}$ & 83.80{\scriptsize$\pm$}$_{0.62}$ & 86.28{\scriptsize$\pm$}$_{5.79}$ & 47.45{\scriptsize$\pm$}$_{1.02}$ & 77.27 & -0.57 \\
    \midrule

    \multirow{5}{*}{\textbf{LLaMA3.1-8B-Instruct}} 
      & \textbf{K16V16}   & \textbf{76.75{\scriptsize$\pm$}$_{1.16}$} & \textbf{47.15{\scriptsize$\pm$}$_{0.48}$} & \textbf{63.96{\scriptsize$\pm$}$_{3.11}$} & \textbf{26.94{\scriptsize$\pm$}$_{5.00}$} & \textbf{53.70} & {-} \\
      & KIVI-K2V2                  & 63.58{\scriptsize$\pm$}$_{0.63}$ & 31.45{\scriptsize$\pm$}$_{0.62}$ & 55.30{\scriptsize$\pm$}$_{8.72}$ & 23.88{\scriptsize$\pm$}$_{8.57}$ & 43.55 & -10.15 \\
      & KIVI-K2V2*           & 71.04{\scriptsize$\pm$}$_{0.83}$ & 44.12{\scriptsize$\pm$}$_{0.62}$ & 56.71{\scriptsize$\pm$}$_{4.88}$ & 23.67{\scriptsize$\pm$}$_{3.88}$ & 48.89 & -4.81 \\
      \rowcolor{lightpink}& \SYS{}         & 75.99{\scriptsize$\pm$}$_{0.96}$ & 45.97{\scriptsize$\pm$}$_{1.57}$ & 60.00{\scriptsize$\pm$}$_{5.85}$ & 25.41{\scriptsize$\pm$}$_{3.47}$ & 51.84 & -1.86 \\
      \rowcolor{lightpink}& \SYS{}-Pro          & 75.51{\scriptsize$\pm$}$_{1.06}$ & 47.37{\scriptsize$\pm$}$_{0.90}$ & 61.65{\scriptsize$\pm$}$_{5.43}$ & 25.82{\scriptsize$\pm$}$_{3.78}$ & 52.59 & -1.11 \\
    \midrule

    \multirow{5}{*}{\textbf{LLaMA3.3-70B-Instruct}} 
      & \textbf{K16V16}   & \textbf{94.92{\scriptsize$\pm$}$_{0.30}$} & \textbf{71.58{\scriptsize$\pm$}$_{1.12}$} & \textbf{83.13{\scriptsize$\pm$}$_{2.03}$} & \textbf{45.92{\scriptsize$\pm$}$_{2.55}$} & \textbf{73.89} & {-} \\
      & KIVI-K2V2                  & 93.91{\scriptsize$\pm$}$_{0.56}$ & 66.05{\scriptsize$\pm$}$_{0.76}$ & 79.07{\scriptsize$\pm$}$_{2.24}$ & 45.07{\scriptsize$\pm$}$_{5.44}$ & 71.03 & -2.86 \\
      & KIVI-K2V2*           & 94.77{\scriptsize$\pm$}$_{0.30}$ & 69.64{\scriptsize$\pm$}$_{0.81}$ & 82.72{\scriptsize$\pm$}$_{2.85}$ & 47.11{\scriptsize$\pm$}$_{2.38}$ & 73.56 & -0.33 \\
      \rowcolor{lightpink}& \SYS{}         & 95.05{\scriptsize$\pm$}$_{0.63}$ & 70.35{\scriptsize$\pm$}$_{0.59}$ & 83.13{\scriptsize$\pm$}$_{1.02}$ & 45.58{\scriptsize$\pm$}$_{3.23}$ & 73.53 & -0.36 \\
      \rowcolor{lightpink}& \SYS{}-Pro          & 95.00{\scriptsize$\pm$}$_{0.23}$ & 70.77{\scriptsize$\pm$}$_{0.59}$ & 82.72{\scriptsize$\pm$}$_{0.81}$ & 46.94{\scriptsize$\pm$}$_{3.06}$ & 73.86 & -0.03 \\
    \bottomrule
\end{tabular}
  \emph{Notes.} 
  \texttt{K16V16} denotes 16-bit precision for the KV cache, while \texttt{KIVI-K2V2} applies 2-bit quantization using the KIVI~\citep{KIVI} algorithm; an asterisk (*) indicates that the first 32 tokens remain in full precision.  
\texttt{\SYS{}} and \texttt{\SYS{}-Pro} preserve 12.5\% and 25\% of channels in INT4.  
The maximum generation length is 4096 tokens.
  
\end{table*}

In order to verify the effectiveness of the quantization scheme presented in Section \ref{sec:algorithm_design}, we conducted comprehensive accuracy comparisons between our methods and existing work.
The overall results for Qwen3 models and LLaMA3 models are summarized in Table \ref{tab:accuracy_overview}.
``Average'' here is the mean across benchmarks, and ``Difference'' is the difference of average accuracy versus the FP16 baseline (K16V16).
For each model, we evaluate the inference accuracy of the baseline KV cache (K16V16), variances of KIVI quantization algorithm, and variances of our \SYS{} algorithm.
The K16V16 here is the original implementation of HuggingFace Tranformers with FP16 KV cache.
KIVI-K4V4 and KIVI-K2V2 use the KIVI~\citep{KIVI} algorithm to quantize the KV cache into 4 and 2 bits.
The details of our \SYS{} variants are described in Section \ref{sec:algorithm_design}, where a small fraction (12.5\%, 25\%) of the key-cache channels are quantized to 4-bit precision while the rest of the channels are quantized to 2-bit precision, results in \SYS{} and \SYS{}-Pro.
Below we summarize the main observations and their practical implications.


\textbf{KIVI-K2V2.}
The KIVI algorithm fails to preserve model accuracy under aggressive 2-bit KV cache quantization.  
As shown in Table~\ref{tab:accuracy_overview}, KIVI-K2V2 consistently yields significantly lower accuracy than the FP16 baseline (\textbf{K16V16}) across all evaluated tasks.  
This degradation becomes particularly evident on reasoning benchmarks such as GSM8K and MATH-Algebra.

\textbf{KIVI-K2V2*.}
KIVI-K2V2* demonstrates substantial accuracy recovery compared to KIVI-K2V2, effectively narrowing the degradation gap across multiple models.  
This improvement aligns with our findings in Section~\ref{sec:design_space_exploration}, highlighting the importance of preserving the initial tokens in full precision.  
However, despite the partial recovery, KIVI-K2V2* still lags behind the FP16 baseline, 
especially on complex reasoning tasks such as MATH-Algebra.

\textbf{\SYS{} and \SYS{}-Pro.}
Our proposed \SYS{} with channel-wise precision boost strategy effectively bridges the remaining accuracy gap left.
As shown in Table~\ref{tab:accuracy_overview}, \SYS{} consistently recovers accuracy across all evaluated models, surpassing KIVI-K2V2* by 5.17 on Qwen3-8B in terms of average accuracy.  
The enhanced variant, \SYS{}-Pro, further increases the boosted channel ratio (from 12.5\% to 25\%), achieving near-parity or even slight improvements over the FP16 baseline on several models such as Qwen3-14B and LLaMA3.3-70B-Instruct.  
These results demonstrate that incorporating channel-wise sensitivity is a principled and scalable approach to restoring accuracy under aggressive 2-bit quantization.

\textbf{Summary.}  
Our results suggest that \emph{channel-wise precision boost} is an effective strategy to mitigate accuracy loss from aggressive KV quantization.  
\SYS{}-Pro achieves near-parity with FP16 accuracy across multiple benchmarks and models, while retaining the memory benefits of low-bit KV cache.

\subsection{Extended Results on Longer Context Length}
In Table~\ref{tab:accuracy_overview}, the maximum generation length is set to 4,096 tokens.  
To examine our method’s robustness under longer-context reasoning, we further evaluated AIME24 and AIME25 with a maximum generation length of 32,768 tokens.  
As shown in Table~\ref{tab:qwen3_8b_aime}, the degradation of 2-bit quantization becomes more pronounced, with KIVI-K2V2 suffering an average accuracy drop of around 13 points compared to the FP16 baseline.  
While KIVI-K2V2* alleviates part of the degradation, it still lags behind the FP16 baseline by 6–8 points.  
In contrast, our proposed \SYS{} achieves substantial recovery, narrowing the average gap to only 3–4 points across all models.  
These results demonstrate that \SYS{} maintains more stable accuracy even at 32k-token context lengths, showing strong robustness under long-context reasoning settings.

\begin{table}[htb]
  \centering
  \caption{Qwen3-8B results on AIME24 and AIME25 (max generation length: 32k tokens).
  KIVI-KV2* here denotes the variant of KIVI where the KVs of initial tokens are not quantized.
  }
  \label{tab:qwen3_8b_aime}
\begin{tabular}{
      l
      l
      c  
      c  
      c  
  }
    \toprule
    \multicolumn{2}{c}{\textbf{Model / Method}} 
      & \multicolumn{1}{c}{\textbf{AIME24}}
      & \multicolumn{1}{c}{\textbf{AIME25}}
      & \multicolumn{1}{c}{\textbf{Avg.}}\\
    \midrule

    \multirow{4}{*}{\rotatebox[origin=c]{90}{\textbf{Qwen-8B}}} 
      & \textbf{KV16}   & \textbf{71.67{\scriptsize$\pm$}$_{15.00}$} & \textbf{66.00{\scriptsize$\pm$}$_{7.33}$} & \textbf{68.84} \\
      & KIVI-KV2                  & 57.00{\scriptsize$\pm$}$_{7.00}$ & 52.33{\scriptsize$\pm$}$_{9.00}$ & 54.67 \\
      & KIVI-KV2*           & 67.67{\scriptsize$\pm$}$_{9.00}$ & 57.67{\scriptsize$\pm$}$_{9.00}$ & 62.67 \\
      & \SYS{}         & 70.67{\scriptsize$\pm$}$_{7.33}$ & 59.67{\scriptsize$\pm$}$_{10.33}$ & 65.17 \\
    \midrule
    \addlinespace[2pt]

    \multirow{4}{*}{\rotatebox[origin=c]{90}{\textbf{Qwen3-14B}}} 
      & \textbf{KV16}   & \textbf{80.67{\scriptsize$\pm$}$_{7.33}$} & \textbf{69.33{\scriptsize$\pm$}$_{7.33}$} & \textbf{75.00} \\
      & KIVI-KV2                  & 69.00{\scriptsize$\pm$}$_{9.00}$ & 55.33{\scriptsize$\pm$}$_{8.67}$ & 62.17 \\
      & KIVI-KV2*           & 74.67{\scriptsize$\pm$}$_{8.00}$ & 61.33{\scriptsize$\pm$}$_{8.67}$ & 68.00 \\
      & \SYS{}         & 75.67{\scriptsize$\pm$}$_{4.33}$ & 67.33{\scriptsize$\pm$}$_{9.33}$ & 71.50 \\
    \midrule
    \addlinespace[2pt]

    \multirow{4}{*}{\rotatebox[origin=c]{90}{\textbf{Qwen3-32B}}} 
      & \textbf{KV16}   & \textbf{81.67{\scriptsize$\pm$}$_{5.00}$} & \textbf{72.59{\scriptsize$\pm$}$_{7.41}$} & \textbf{77.13} \\
      & KIVI-KV2                  & 73.00{\scriptsize$\pm$}$_{9.67}$ & 57.41{\scriptsize$\pm$}$_{9.26}$ & 65.21 \\
      & KIVI-KV2*           & 79.00{\scriptsize$\pm$}$_{9.00}$ & 59.05{\scriptsize$\pm$}$_{12.38}$ & 69.03 \\
      & \SYS{}         & 79.67{\scriptsize$\pm$}$_{9.67}$ & 69.26{\scriptsize$\pm$}$_{9.26}$ & 74.47 \\
    \bottomrule
\end{tabular}
\end{table}

\subsection{Ablation Study on Channel-wise Precision Boost}
\label{sec: promotion_rate}

\begin{figure}[htb]
    \centering
    \subfloat[GSM8K]{
        \includegraphics[width=0.23\textwidth]{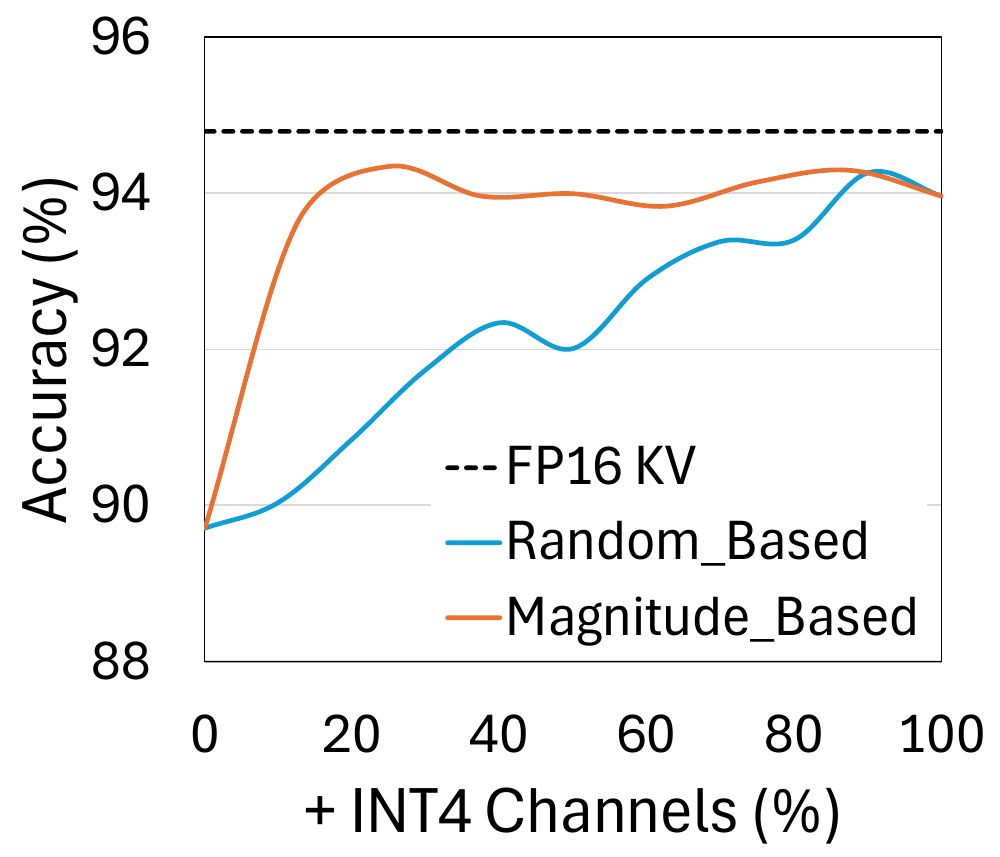}
    }
    \hfill
    \subfloat[MATH-Algebra]{
        \includegraphics[width=0.23\textwidth]{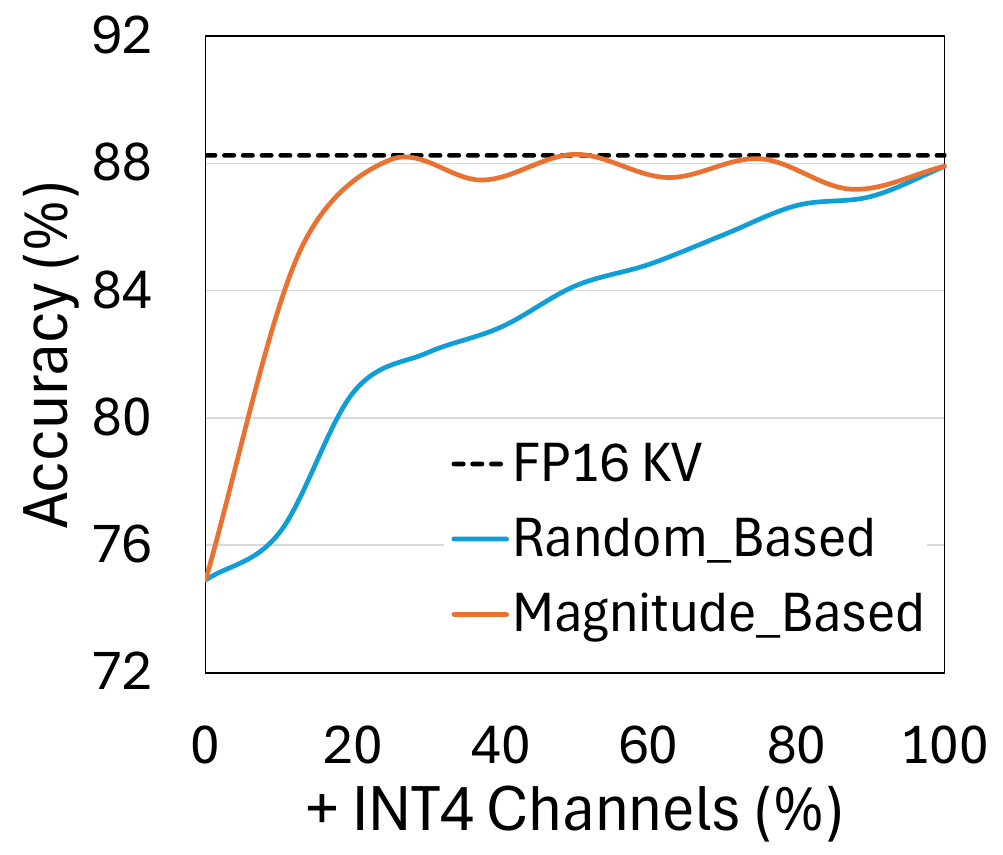}
    } \\
    \caption{Accuracy recovery with \emph{channel-wise precision boost} on Qwen3-8B.
    Similar trends are observed on other tasks.}
    \label{fig:qwen3-8b-ablation}
\end{figure}

In Section \ref{sec:accuracy_overview}, we demonstrate the effectiveness of \SYS{} and \SYS{}-Pro.
In this section, we conduct ablation study by more comprehensive investigating the impact of using different \emph{channel boost rate}\footnote{We use the term \emph{channel boost rate} to denote the fraction of channels boosted from 2-bit to 4-bit precision.}.
Figure~\ref{fig:qwen3-8b-ablation} shows the accuracy trends on Qwen3-8B when varying the fraction of key-cache channels boosted to INT4.
A flat baselines are also plotted for K16V16 baseline (upper dashed line).

\textbf{Monotonic Accuracy Improvements.}
As shown in Figures \ref{fig:qwen3-8b-ablation}, accuracy on GSM8K and MATH\_Algebra improves almost monotonically as the \emph{channel boost rate} increases.
Notably, accuracy is effectively recovered once 25\% of channels are boosted to 4-bit precision.
This mirrors the table-level averages in Table~\ref{tab:accuracy_overview}, where \SYS{}-Pro reaches parity.
Overall, we find that accuracy improves consistently with higher \emph{channel boost rates} across most tasks.

\textbf{Necessity of Selection Heuristic.}
To examine whether heuristics are indeed necessary for channel-wise precision boost, we also include a random selection baseline, where each channel is assigned a random importance score.
By comparing the accuracy curves in Figures~\ref{fig:qwen3-8b-ablation}, we verified that while randomly boost some channels can improve accuracy, heuristic-guided \emph{channel-wise precision boost} yields substantially greater benefits.

\subsection{System Results: Inference Efficiency}

\begin{figure}[htb]
    \centering
    \subfloat[GPU Memory Usage.\label{fig:MemoryUsage}]{
        \includegraphics[width=0.35\textwidth]{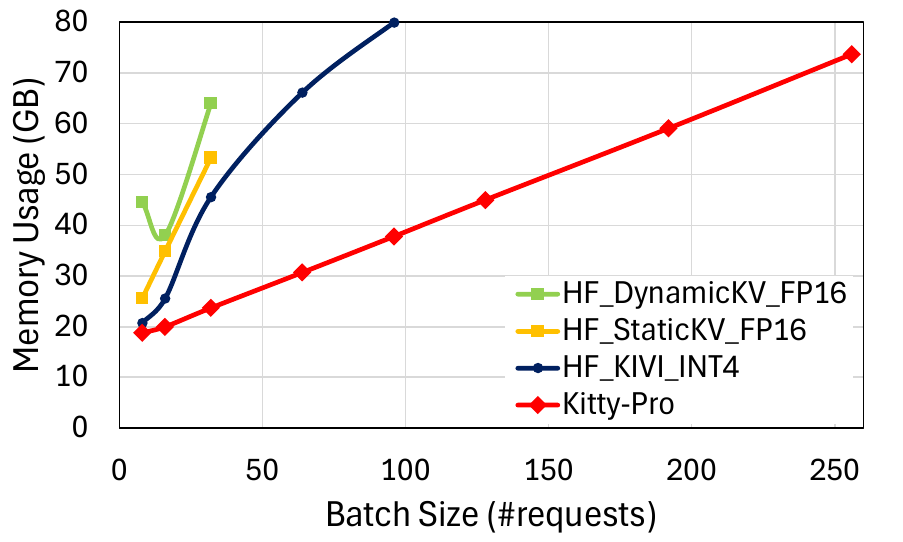}
    }
    \\

    \subfloat[Inference Throughput. \label{fig:TokenThroughput}]{
        \includegraphics[width=0.35\textwidth]{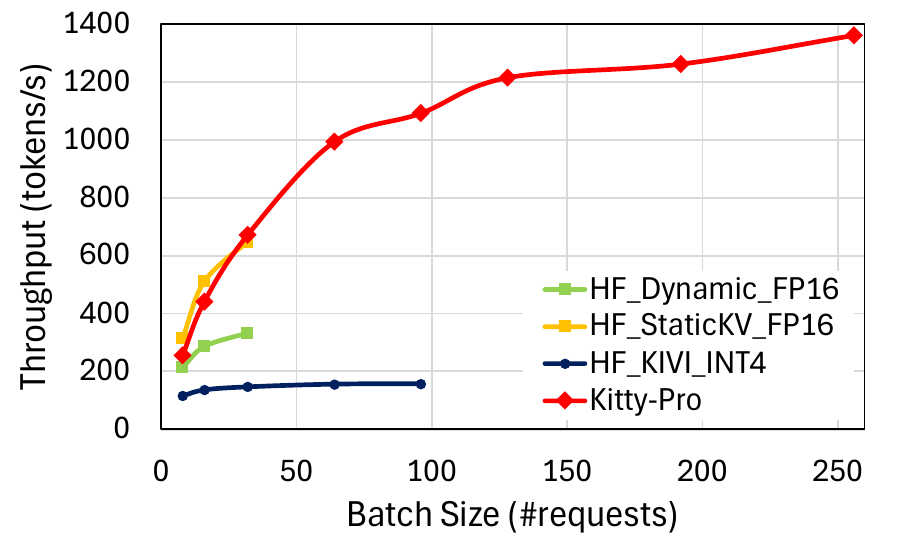}
    }

    \caption{Memory usage and throughput comparison on Qwen3-8B generating 8192 tokens.
    \SYS{} can achieve higher throughput by enabling larger batch sizes.}
    \label{fig:Inference_Efficiency}
\end{figure}

To evaluate system-level efficiency of \SYS{}, we performs end-to-end Qwen3-8B inference using our prototype inference engine presented in Section \ref{sec:system_design}.
A batch of example prompts are fed to the inference system and we let the system generate tokens until it hits the predefined maximum sequence length (8192).
We also include two Hugging Face baselines here to demonstrate the system-level benefits of our KV quantization method.
\emph{HF\_Dynamic\_FP16} here denotes the default configuration of Hugging Face transformers library, where the KV cache grows dynamically as more KV vectors are cached in the system.
\emph{HF\_Static\_FP16} is a more runtime-efficient KV cache implementation, where the KV cache is pre-allocated before the token generation.
Furthermore, we also evaluated the INT4 KV cache implemented by Hugging Face, denoted as HF\_KIVI\_INT4, which is a re-implementation of KIVI~\citep{KIVI}.
The prompts are around 100 tokens so the inference time is mainly dominated by the decoding phase.
For each KV cache implementations, we increase the batch size until out-of-memory, and report the peak memory usage and token generation throughput under different batch sizes.
The hardware we used is a single NVIDIA A100~\citep{Ampere_WhitePaper} GPU (80GB).

As shown in Figure \ref{fig:Inference_Efficiency}, with the same memory budget, \SYS{}-Pro enables $8\times$ batch sizes and achieves $2.1\times \to 4.1\times$ higher inference throughput compared to the FP16 baseline.
We also note that we implemented our customized GPU kernels with Triton~\citep{Triton} for proof-of-concept.
Thus, the inference efficiency of our inference engine can be further improved if we use more low-level programming language, e.g., CUDA, and enable more fine-grained optimizations.
We leave as one of future work.

\section{Conclusions}
\label{sec:conclusion}
This paper identified that 4-bit KV cache quantization largely preserves accuracy, while existing 2-bit quantization method leads to substantial degradation on reasoning-intensive tasks.
Preserving initial tokens in full precision helps but does not fully close the gap with FP16. 
Our channel-wise precision boost method further narrows this gap: boosting only 12.5\%-25\% of key-cache channels to higher precision is often sufficient to recover most of the accuracy loss.
Based on this innovation, we propose an accurate and memory-efficient quantization framework, \emph{\SYS{}}.
Furthermore, we propose corresponding system designs to support the end-to-end inference with \SYS{}.
The end-to-end evaluations show that our 2-bit inference system can significantly reduce the GPU memory consumption and enable $8\times$ larger batch sizes during inference, which results in $2.1\times \to 4.1\times$ higher inference throughput compared to the FP16 baseline with same memory budget.





\bibliography{main}
\bibliographystyle{mlsys2025}



\end{document}